%% file: root.tex
\documentclass[letterpaper, 10 pt, conference]{ieeeconf}  
\usepackage{hhline}
\usepackage{epsfig}
\usepackage{epstopdf}
\usepackage{graphics}
\usepackage{epstopdf}
\usepackage{upgreek}
\usepackage{amssymb}
\usepackage{amsfonts}
\usepackage{amsmath}
\usepackage{refstyle}
\usepackage{xcolor}
\usepackage{subfigure}
\usepackage{todonotes}
\usepackage{mathtools}
\usepackage{algorithm}
\usepackage{algpseudocode}
\usepackage{cite}
\usepackage{cleveref}
\usepackage{url}
\usepackage{leftidx}
\usepackage[detect-all]{siunitx}
\usepackage{makecell}
\usepackage{multirow}
\usepackage{float}
\usepackage{makecell}
\usepackage{booktabs}
\usepackage{wrapfig}
\usepackage{tablefootnote}
\usepackage{threeparttable}

\UseRawInputEncoding
\usepackage{balance}
\usepackage[font=footnotesize]{caption}
\usepackage[export]{adjustbox}

\makeatletter 
\newenvironment{mcases}[1][l]
 {\let\@ifnextchar\new@ifnextchar
  \left\lbrace
  \array{@{}l@{\quad}#1@{}}}
 {\endarray\right.}
\makeatother

\makeatletter
\newcommand*{\centerfloat}{%
  \parindent \z@
  \leftskip \z@ \@plus 1fil \@minus \textwidth
  \rightskip\leftskip
  \parfillskip \z@skip}
\makeatother

\IEEEoverridecommandlockouts                              

\title{\LARGE \bf Sim-to-Real Transfer of Robotic Assembly with Visual Inputs \\Using CycleGAN and Force Control}

\author{Chengjie Yuan$^{1,2 \, \dagger}$, Yunlei Shi$^{3,2 \, \dagger}$, Qian Feng$^{1,2}$, Chunyang Chang$^{2}$,\\ Zhaopeng Chen$^{2}$, Alois Christian Knoll$^{1}$, Jianwei Zhang$^{3}$
\thanks{*This research has received funding from the German Research Foundation (DFG) and the National Science Foundation of China (NSFC) in project Crossmodal Learning, DFG TRR-169/NSFC 61621136008, partially supported by the European Union’s Horizon 2020 research and innovation
programme under the Marie Sklodowska-Curie grant agreement No 691154 STEP2DYNA and No 778602 ULTRACEPT.}
\thanks{{$^{1}$Technische Universit\"at M\"unchen}, {$^{2}$Agile Robots AG}, {$^{3}$TAMS (Technical Aspects of Multimodal Systems), Department of Informatics, Universit\"at Hamburg}}
\thanks{$^{\dagger}$ The first two authors contributed equally to this work.}
}

\DeclarePairedDelimiterX{\norm}[1]{\lVert}{\rVert}{#1}
\begin{document}
\addtolength{\topmargin}{0.08in}

\makeatletter
\let\@oldmaketitle\@maketitle
\renewcommand{\@maketitle}{\@oldmaketitle
  \includegraphics[width=\linewidth]
    {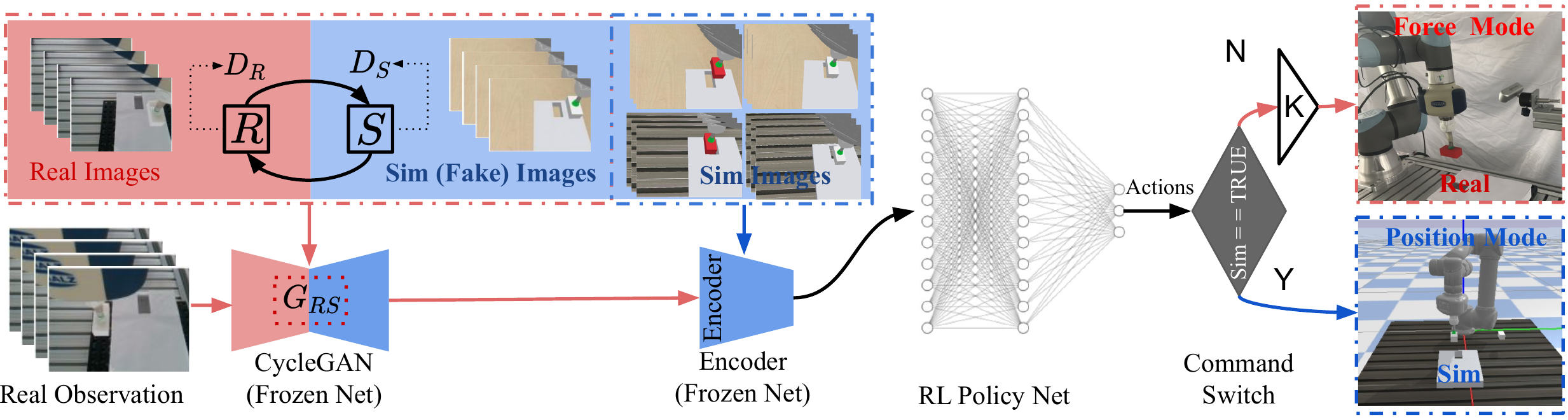} \\[0.25em]
  \refstepcounter{figure}\footnotesize{Fig. 1: Sim-to-real learning-based framework for a rectangular peg-in-hole insertion task. Sim part (in blue) is used to train the encoder (Frozen Net) and reinforcement learning (RL) policy net in a simulator. $G_{RS}: R \rightarrow S$ is a mapping function generated using a cycle-consistent generative adversarial networks (CycleGAN) to transfer an image from a real world style to a simulator style. A command switch is used to ensure safety in the contact-rich operation by changing the position control mode in the simulator to a force control mode in the real world.}
  \label{fig:real} \medskip \vspace{-10pt}}
\makeatother

\maketitle
\thispagestyle{empty}
\pagestyle{empty}

\newcommand{\hand}{DLR-HIT Hand II}

\input{chapters/00_abstract.tex}

\input{chapters/01_introduction.tex}

\input{chapters/02_related_work.tex}

\input{chapters/03_problem_statement.tex}

\input{chapters/04_framework_design.tex}

\input{chapters/05_experiments.tex}

\input{chapters/06_conclusion.tex}



\linespread{1.0}
\bibliographystyle{IEEEtran}
{\scriptsize
\vspace{0.01 cm}
\bibliography{references}
}

\end{document}

%% file: chapters/00_abstract.tex
\begin{abstract}
Recently, deep reinforcement learning (RL) has shown some impressive successes in robotic manipulation applications. However, training robots in the real world is nontrivial owing to sample efficiency and safety concerns. Sim-to-real transfer is proposed to address the aforementioned concerns but introduces a new issue called the reality gap. In this work, we introduce a sim-to-real learning framework for vision-based assembly tasks and perform training in a simulated environment by employing inputs from a single camera to address the aforementioned issues. We present a domain adaptation method based on cycle-consistent generative adversarial networks (CycleGAN) and a force control transfer approach to bridge the reality gap. We demonstrate that the proposed framework trained in a simulated environment can be successfully transferred to a real peg-in-hole setup.
\end{abstract}

%% file: chapters/01_introduction.tex
\section{INTRODUCTION}
\textcolor{black}{Industrial robots are commonly used in structured environments, such as car manufacturing factories and phone assembly lines. The requirement to push the border of the \textcolor{black}{"Robot Zone"} \cite{Filho2016Automatic} toward the manual manufacturing domain is increasing rapidly. \textcolor{black}{Humans can execute manual manufacturing tasks easily using visual and force feedback, whereas robotic conventional methods, such as position control or visual servoing, are difficult to accomplish.} Reinforcement learning (RL) shows potential to solve complex robot manipulation problems because it allows an agent to interact with the environment for trial-and-error learning and accepts high-dimension feedback as the input.\cite{lee2019making}, \cite{sutton2018reinforcement}, \cite{kober2013reinforcement}.}

\textcolor{black}{For contact-rich manipulations, \textcolor{black}{it is nontrivial to establish a robotic system that can learn a task with a safety guarantee and avoid wear and tear problem.}} Thus, \textcolor{black}{sim-to-real methods are proposed \cite{peng2018sim} to address the aforementioned concerns}. Recently, style transfer methods based on generative adversarial networks (GANs) \cite{goodfellow2014generative} have been proposed recently in the computer vision field, enabling the use of vision-based  manipulation tasks for deploying visual sim-to-real methods; however, owing to poorly simulated dynamics, the sim-to-real reality gap could be an issue when transferred \textcolor{black}{the simulated policies} to physical setups \cite{rao2020rl}.

In this work, we verify \textcolor{black}{our framework} using the the most commonly used assembly task: peg-in-hole \cite{park2013intuitive}. The framework training is performed in a simulated environment by employing only images captured using a camera as the input. When transferring the trained policy in our framework to a physical robot, the execution command is mapped from the position signal to the force signal to assist the peg-in-hole insertion task (we employ an admittance controller to perform the compliant movement).

Our framework can be described as follow. First, we train the RL \textcolor{black}{policy net (we use soft actor–-critic (SAC) algorithm in this work)} in the simulator. Then, we use the images of insertion scenarios collected from a simulator and the real world to train a cycle-consistent generative adversarial networks (CycleGAN) \cite{zhu2017unpaired}. Thereafter, the trained policy is driven by the real-to-sim transformed image style, obtained from the trained CycleGAN. The entire framework is shown in Figure 1.\\ 

Our primary contributions are listed:

\begin{itemize}
\item \textbf{C1:} A vision-based sim-to-real learning framework is proposed to perform assembly tasks.

\item \textbf{C2:} A peg-in-hole task that effectively leverages visual information and force control using a simple reward function for a complete insertion, including hole searching, alignment, and insertion. The task performance is compared when training using different visual observation spaces.

\item \textbf{C3:} The robustness of the framework to perturbations and sensor noise in real world is evaluated.
\end{itemize}

The remainder of the paper is structured as follows. In \Cref{RELATED WORK AND BACKGROUND}, we describe the background and development of the deployed method. \Cref{PROBLEM STATEMENT} introduces the problems. In \cref{POLICY DESIGN}, we provide an overview of our method. A quantitative experiment of our methods and the experiment results in \Cref{EXPERIMENTS}. \Cref{RESULTS AND DISCUSSION} presents the conclusions and future work. \\

%% file: chapters/02_related_work.tex
\section{RELATED WORK AND BACKGROUND}\label{RELATED WORK AND BACKGROUND}
\subsection{Contact-Rich Assembly}\label{Assembly strategies}

The entire process of an assembly task can be considered as a constrained motion with geometrical and environmental constraints. Generally, we can decompose the existing peg-in-hole assembly strategies into two categories: contact model-based and contact model-free strategies \cite{xu2019compare}. Model-based strategies rely on the contact model analysis and compliant control. Two common examples of model-free strategies are learning from demonstration (LFD) and RL. 
For contact model analysis, analytical and statistical models are commonly used \cite{1982Quasi}, \cite{2012Contact}, \cite{1998Contact}, \cite{2002A}, \cite{2014Contact}, \cite{hovland1998hidden}. Analytical models are based on the analysis of geometrical and environmental constraints, whereas statistical models rely on the estimation of the contact state by collecting samples. Among contact model-free strategies, the LFD approaches can be categorized into three phases: sensing, encoding, and reproducing \cite{zhu2018robot}, \cite{kyrarini2019robot}. However, the sample efficiency of the aforementioned methods highly relies on the human operation and these methods introduce transparency problems in contact-rich teaching tasks \cite{shi2021combining}.

\subsection{Reinforcement Learning}\label{reinforcement learning}

The RL is a machine learning approach for teaching agents to solve different tasks based on trials and errors when interacting with environments. The RL agent aims to learn a policy $\pi(a_{t}|s_{t})$, which selects the action $a_{t}$, and meanwhile the agent observes the environment $s_{t}$. The transition probability $p(s_{t+1}|a_{t},s_{t})$ is used to connect the state change over dynamics. The final trajectory can be represented as $\tau = (s_{0},a_{0},s_{1},a_{1},...)$. The discount factor $\gamma$ controls the sum of the reward.
An optimal policy $\pi^{*}$ should maximize the cumulative reward $r(s_{t},a_{t})$ during interactions with the environment, as shown in \Cref{equ: optimal policy}.

\begin{equation}\label{equ: optimal policy}
    \pi^{*} =  \operatorname*{arg\,max}_{\pi} \mathbb{E}_{\tau \sim \pi}\bigg[\sum_{t=0}^{\infty}\gamma^{t}(r(s_{t},a_{t}))\bigg]
\end{equation}

With the development of expressive function approximation such as neural networks (e.g., Deep RL), high-dimensional inputs such as raw images can be handled \cite{mnih2013playing}, \cite{sunderhauf2018limits}. Great success has been gained because of the advances in RL in many fields, for instance, the development of video games such as Atari \cite{mnih2013playing}, dexterous hand manipulation \cite{andrychowicz2020learning}, robot grasping \cite{kalashnikov2018qt}, and robot manipulation \cite{levine2016end}. 

RL algorithms can be classified into two branches: model-based and model-free algorithms \cite{sutton2018reinforcement}. The main difference between the two branches is whether an agent gets access to or learns a model of the environment.
Different assembly tasks are also solved using a model-based RL called mirror descent guided policy search (MDGPS) \cite{luo2018deep}. By combining the force information obtained from a force-torque sensor at the end-effector with a long short-term memory (LSTM), a high-precision assembly task was performed \cite{inoue2017deep}. An operational space control framework was used in \cite{kaspar2020sim2real}. By combining visual inputs with natural rewards, different connector insertion tasks were demonstrated \cite{schoettler2019deep}. InsertionNet \cite{spector2021insertionnet} was proposed to solve the general insertion problem by combining visual and force inputs, and it was trained in the real-word environment, which has safety risks \cite{hofer2020perspectives}.

\subsection{CycleGAN} \label{CycleGAN}
CycleGAN is an extension of GANS \cite{goodfellow2014generative}. GANs comprise two submodels: generator and discriminator models. The key idea of GANs is to train the two submodels in a zero-sum adversarial game, until the discriminator is fooled by the generated examples half the time. In CycleGAN, except for the original adversarial loss, a cycle consistency loss is proposed: this loss can be used to calculate the reconstruction error of the images. Furthermore, CycleGAN offers a considerably more efficient approach for training than common GANs because of using unpaired and unlabeled dataset.

\subsection{Sim-to-Real Transfer}\label{Sim-to-Real Transfer}
Training in a simulator can assuredly provide an infinite amount of data and alleviate certain safety concerns during training. However, a reality gap exists that often separates a simulated task and its real-world analog, leading to failure when working with physical robots. To bridge the reality gap, three main options for solving this problem: system identification (SI), domain adaptation (DA) and domain randomization (DR) \cite{ibarz2021train}. Using SI, we can remove inaccessible states and apply state estimation during the training \cite{tan2018sim}. DR helps the trained policy to adapt to different dynamics and generalize to dynamics of the real world \cite{peng2018sim}, allowing the randomization of parameters for visual or pixel-based inputs such as lighting and textures \cite{tobin2017domain}.

For dexterous hand manipulation, a trained policy combing both dynamic and visual randomization is deployed \cite{andrychowicz2020learning}.\begin{figure}[!h] 
\addtocounter{figure}{-1}
\centering
\subfigure[Simulator]{\includegraphics[width=2.8cm,height=4cm]{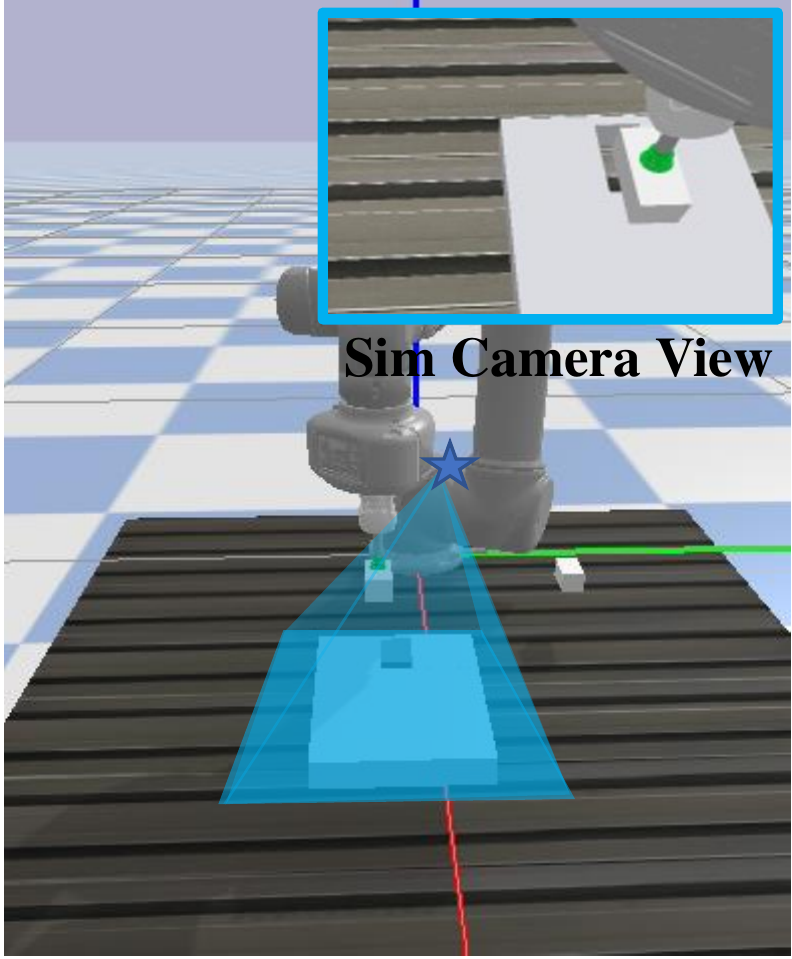}}
\subfigure[Real-world setup]{\includegraphics[width=5.7cm,height=4cm]{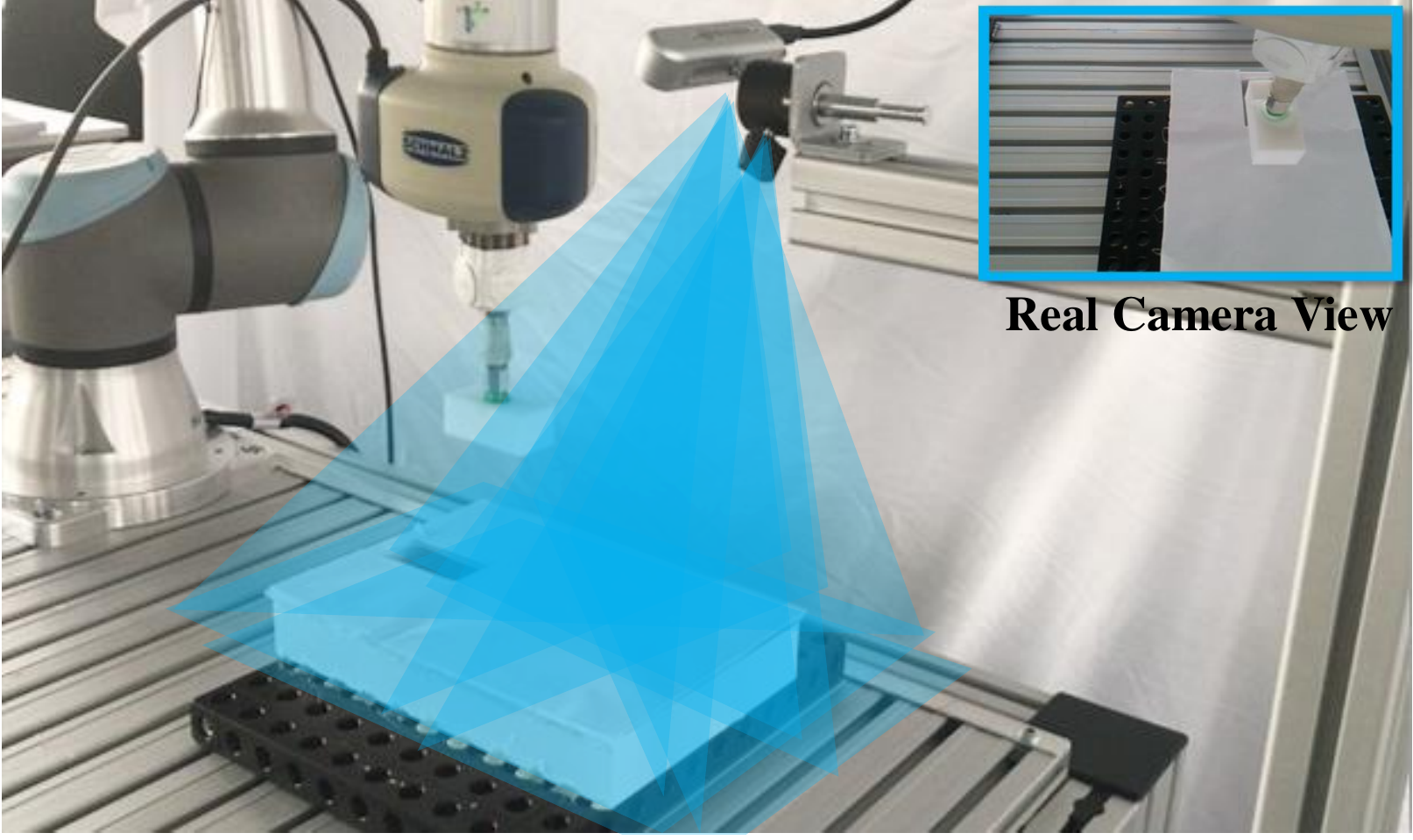}}
\caption{\textcolor{black}{Simulator and real-world setup: The blue area represents the view of the camera and the scene in camera can be seen in the insets of the two images (a) and (b).\\}} 
\label{fig:Simulated and real setup}
\end{figure}
The DA approach aims to map the source domain to the target domain; in the robotic context, this approach usually exploits the recent advances in visual domain adaptation \cite{2017Unsupervised}, \cite{arndt2020meta}, \cite{james2019sim}.

%% file: chapters/03_problem_statement.tex
\section{PROBLEM STATEMENT}\label{PROBLEM STATEMENT}

Owing to unknown contact mechanics, \textcolor{black}{designing a feedback control mechanism for contact-rich tasks is challenging. RL has shown some progress in robotic contact-rich tasks in unstructured environments}; however, sample efficiency and safety concerns are two main problems when performing policy training. Many RL algorithms require millions of steps to train policies \textcolor{black}{for performing} complex tasks \cite{levine2018learning}, \cite{lee2019making}.  In other words, human supervision is always needed in \textcolor{black}{resetting experiments, hardware status monitoring, and safety assurance,} which is quite time-consuming and tedious \cite{ibarz2021train}.

The sim-to-real approach shows potential to solve the aforementioned problems; however, one significant difficulty associated with this approach is bridging the reality gap to address the mismatch in distinct distributions of rendered images and real-world counterparts. Another challenge is ascribed to force modeling in \textcolor{black}{simulation as the force interactions} will inevitably occur between the target object and environments when performing contact-rich tasks. Moreover, \textcolor{black}{it is expensive to apply the system calibration due to the limitation of the simulation domain expert's ability} \cite{spector2021insertionnet} and accurate requirements \cite{hofer2020perspectives}.

%% file: chapters/04_framework_design.tex
\section{FRAMEWORK DESIGN}\label{POLICY DESIGN}

In this section, we present the overview of our sim-to-real framework proposed to solve the problems stated in \Cref{PROBLEM STATEMENT}. We built our system in a Bullet simulator \cite{bullet} and modeled the robot and task environment based on OpenAI Gym \cite{brockman2016openai}. \Cref{fig:Simulated and real setup}(a) and \Cref{fig:Simulated and real setup}(b) shows the setup in the simulator and  real world, respectively. 

\begin{figure}[!h]
\centering
\subfigure[]{\includegraphics[width=2.8cm,height=2.8cm]{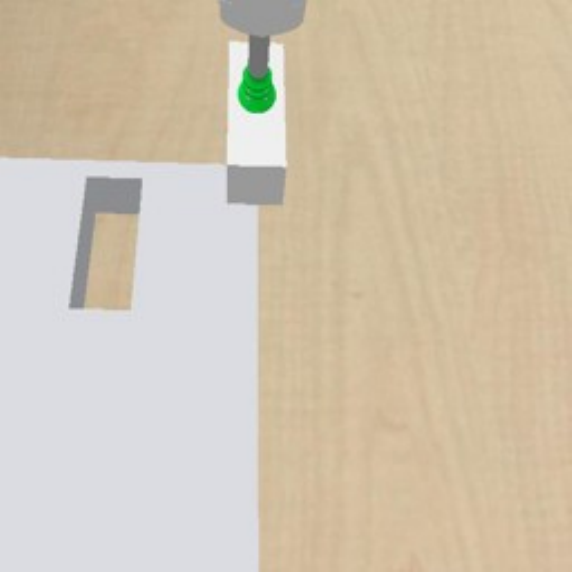}}
\subfigure[]{\includegraphics[width=2.8cm,height=2.8cm]{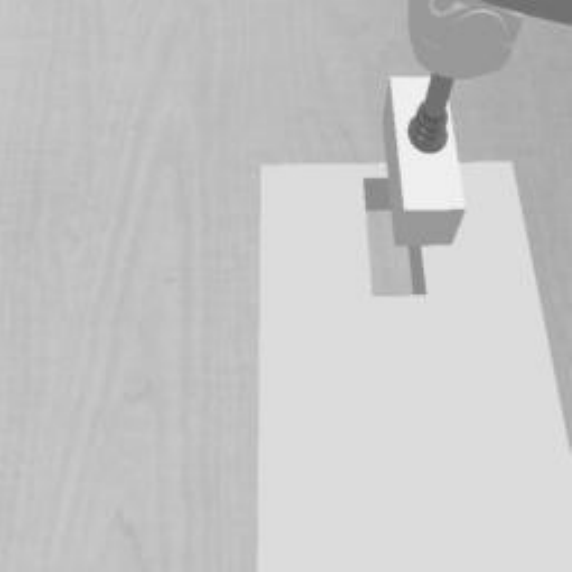}}
\subfigure[]{\includegraphics[width=2.8cm,height=2.8cm]{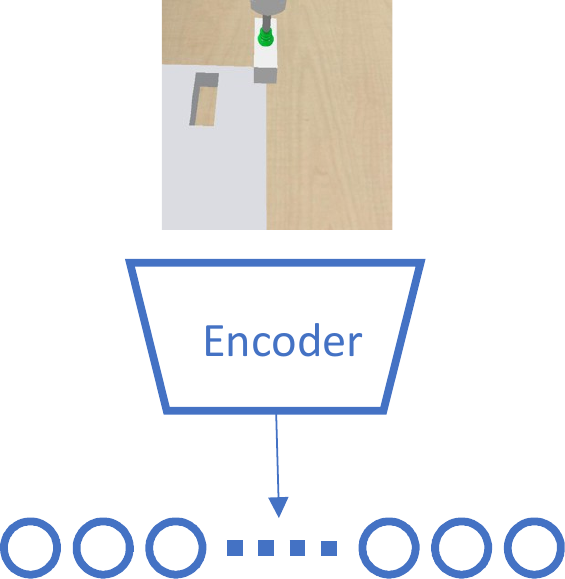}}
\caption{\textcolor{black}{ Three different observation spaces: (a) a raw RGB image, (b) grayscale image, and (c) latent representation generated by an encoder.}}%
\label{fig:Three different observation spaces}
\end{figure}

\subsection{Learning Framework in Simulation}\label{Learning Framework in Simulation}
\vspace{1ex}
\subsubsection{Policy}
The soft actor–-critic (SAC) algorithm \cite{haarnoja2018soft} was employed in our framework. SAC introduces an entropy $H$ in its objective function (\Cref{equ: sac}), which is a significant characteristic, where $\alpha$ denotes a temperature parameter that determines the importance of the entropy term.
The entropy is used to measure the randomness of a given policy. In this study, the policy is trained to maximize \textcolor{black}{a value that relies on the expected return value as well as the entropy. It helps to reach a good trade--off between exploration and exploitation}.

\begin{equation}\label{equ: sac}
    \pi^{*} =  \operatorname*{arg\,max}_{\pi} \mathbb{E}_{\tau \sim \pi}\bigg[\sum_{t=0}^{\infty}\gamma^{t}(r(s_{t},a_{t})+\alpha H(\pi(\cdot|s_{t})))\bigg]
\end{equation}

\vspace{1ex}
\subsubsection{States}

For a vision-based learning policy, the commonly used observation states are the RGB, grayscale, and latent representation \cite{lee2019making}, \cite{schoettler2019deep}. In this work, we select observation spaces as follow:
\begin{itemize}
    \item RGB observation space: $3\times 64\times 64$ tensor
    \item Grayscale observation space : $1\times 64\times 64$ tensor
    \item Latent representation observation space: $128\times 1$ vector.
\end{itemize}

For the RGB and grayscale observation spaces, the network conducts end-to-end learning; in other words, \textcolor{black}{raw images are inputted to the network and the output command is obtained}. For the latent representation observation space, an autoencoder is employed as a part of the network. This autoencoder comprises an encoder and a decoder, we exploit the encoder to compress the input image and generate the latent representation observation space.

\subsubsection{Actions}
Inspired by the literature \cite{lee2018making}, the necessary translation movement along the X-, Y-, and Z axes are considered and the orientation of the end-effector is fixed. We define a three-dimensional (3D) vector that contains the translation movement information of the robot. We use a position controller in the simulation, and the robot will move along a relative distance with respect to the current pose. 
The continuous 3D displacement action space $\Delta P$ 
\begin{align} \label{equ: deltaP}
      \Delta P= [\Delta x,\ \Delta y,\ \Delta z],
\end{align}  
which considers translation movement along the X-, Y-, and Z-axes. The value in each axis is strictly in the interval of $[-0.02,0.02]$ m.

\vspace{1ex}
\subsubsection{Rewards}
Some researchers set reward functions based on the different insertion phases such as reaching, alignment and insertion \cite{lee2018making}, \cite{inoue2017deep}, making the reward function hard to design; and need to distinguish the different phases.
We only design one normal reward function that combines L1 and L2 distances for reaching, alignment and insertion phases and one reward for successfully insertion:

\[
R(\mathbf{s}) =
  \begin{mcases}[l@{\ }]
      50,  & \text{(Success)} \\
     -(flag*10 + 0.4*(\left \| p_{obj}-p_{goal}\right\|) \\
     +0.6*(|p_{obj}-p_{goal}|)) &\text{(Otherwise)},
  \end{mcases}
\]

where $p_{obj}$ and $p_{goal}$ represent the positions of the peg and hole, respectively, and \textit{flag} is set to 1 if the robot moves to a distance exceeding a certain threshold (i.e., 15 cm away from the hole center); otherwise, it is set to 0. Here, \textit{flag} works as a punishment when the robot makes unexpected movements.

\subsection{Transfer Framework to Real-world Environment}\label{Transfer Framework to Real Environment}

\subsubsection{Observation Space Transfer}
To transfer our policy from the simulator to the real world, we must transfer the images from the domain of the real world to its counterparts in the simulator. Conventionally, training an image-to-image translation model requires a paired dataset. The requirement for paired examples is a limitation, it is challenging and expensive to prepare these datasets.

A successful approach for unpaired image-to-image translation is the CycleGAN. We commanded the robot to move randomly in the view of the camera and captured its random state each time. Approximately 200-300 images can be effortlessly obtained for training the model. Using the style transfer based on the CycleGAN, we map the view of the camera in the real-world environment to its counterpart, which we use to train our policy.

\vspace{1ex}
\subsubsection{Action Space Transfer}

In a study \cite{spector2021insertionnet}, researchers incorporated force augmentations by multiplying a random constant $\alpha$ with the force and the moment because they arrived at the conclusion that \textbf{the direction of the vectors $(F,M)$}, but not the magnitude, \textbf{is the most important factor} in insertion operations. We extended this conclusion to our sim-to-real transfer process using a new method: we multiply gain $K$ and the original position action output $ \Delta P$ and then use the product $C_{real}$ as the control command for the real robot force controller:
\begin{equation}
\begin{split}
\boldsymbol{C}_{real} & = [F_{x},F_{y},F_{z}] \\
& = K \Delta P \\
& = K[\Delta x,\ \Delta y,\ \Delta z],
\label{equ: F scale}
\end{split}
\end{equation}
where $K=100$ N/m; thus, the force command values along the X-, Y-, and Z-axes are in the range $[-2,2]$ N.

%% file: chapters/05_experiments.tex
\section{EXPERIMENTS}\label{EXPERIMENTS}

In this section, we introduce the peg-in-hole insertion task to validate our framework and explain the experimental results for both the simulated and real-world environments, in which we address the following questions: 

\begin{figure}[!h]
\centerline{\includegraphics[width=8.7cm,height=6.7cm]{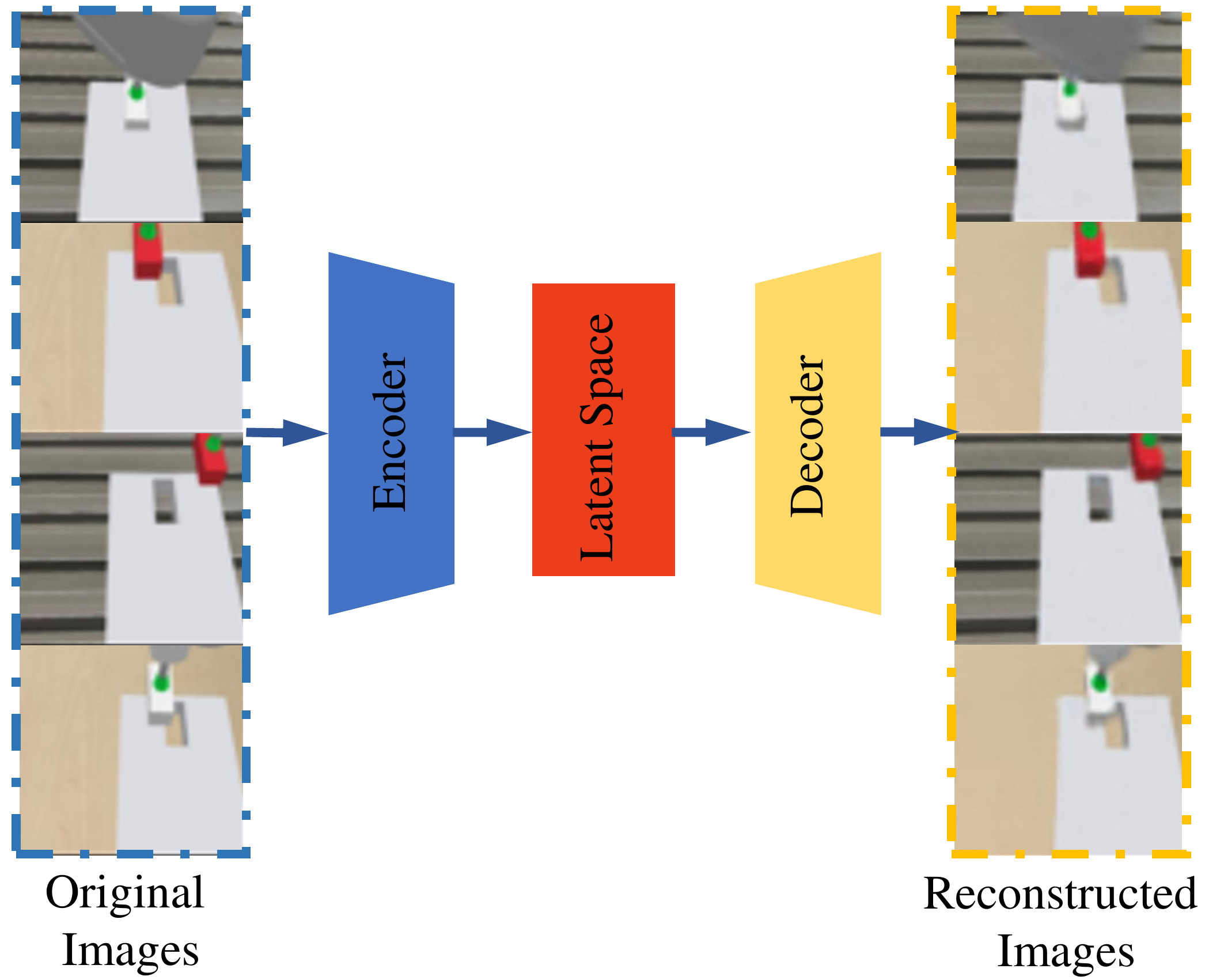}}
\caption{\textcolor{black}{Results obtained using an autoencoder: where the left panels show the original images and the right panels show the reconstructed images.}}
\label{fig: result of autoencoder}
\end{figure}

\begin{table}[h!]
\caption{SUCCESS RATES OF 3 OBSERVATION SPACES}
\label{tab:Performance with different observation space}
\centering
\begin{tabular}{@{}|c|c|c|@{}}
\toprule
\multicolumn{1}{|l|}{\textbf{\quad Total episodes \quad}} & \multicolumn{1}{l|}{\textbf{ \quad Observation space \quad}} & \textbf{\quad Success rate \quad} \\ \midrule
                                              & Gray $1 \times64 \times 64$                             & 0\%                   \\ \cmidrule(l){2-3} 
                                              & RGB $3 \times 64 \times 64$                              & 0\%                   \\ \cmidrule(l){2-3} 
\multirow{-3}{*}{3000}                        & \textbf{Latent $128 \times 1$}                             & \textbf{96\%}         \\ \midrule
                                              & Gray $1 \times 64 \times 64$                             & 0\%                   \\ \cmidrule(l){2-3} 
                                              & RGB $3 \times 64 \times 64$                              & 0\%                   \\ \cmidrule(l){2-3} 
\multirow{-3}{*}{10000}                       & {\color[HTML]{333333} \textbf{Latent $128 \times 1$}}      & \textbf{96\%}         \\ \bottomrule
\end{tabular}
\end{table}

\begin{enumerate}
    \item Will all observation spaces work well in our framework?
    \item Can our trained policy be transferred to a real-world environment successfully?
    \item How does our framework perform compared with other insertion methods in terms of the success rate?
    \item What is the robustness of our framework under external perturbations and target uncertainties?
\end{enumerate}

\subsection{Simulation Experiment}

\subsubsection{Observation Space Comparison}
To compare the performances of the policy with different observation spaces, we test the scene of a white block with a metallic texture. We use the success rate of a complete insertion task in the simulation as a criterion to evaluate the performance of the learned policy. 

A convolutional neural network is utilized as a part of the SAC network for training using the inputs from the RGB and grayscale observation spaces. An autoencoder is used to obtain a latent representation of the input image. The encoder part allows the compression of the original image to a lower-dimension vector that contains the important information.
\begin{figure}[!h]
\centerline{\includegraphics[width=8.5cm,height=2.0cm]{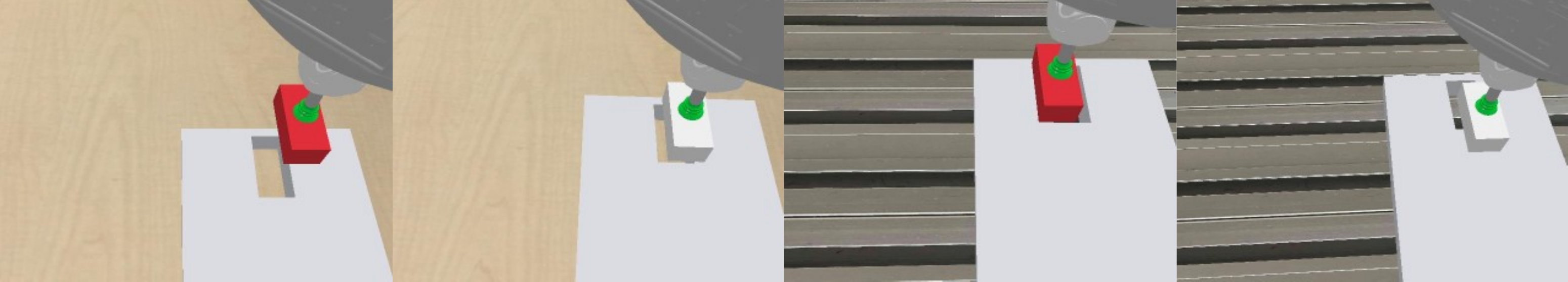}}
\caption{\textcolor{black}{ Four environmental scenes. From left to right: a red block with a wooden texture, a white block with a wooden texture, a red block with a metallic texture, and a white block with a metallic texture.}} 
\label{fig: fourscene}
\end{figure}
\begin{figure}[!h]
\centerline{\includegraphics[width=8.5cm,height=4.5cm]{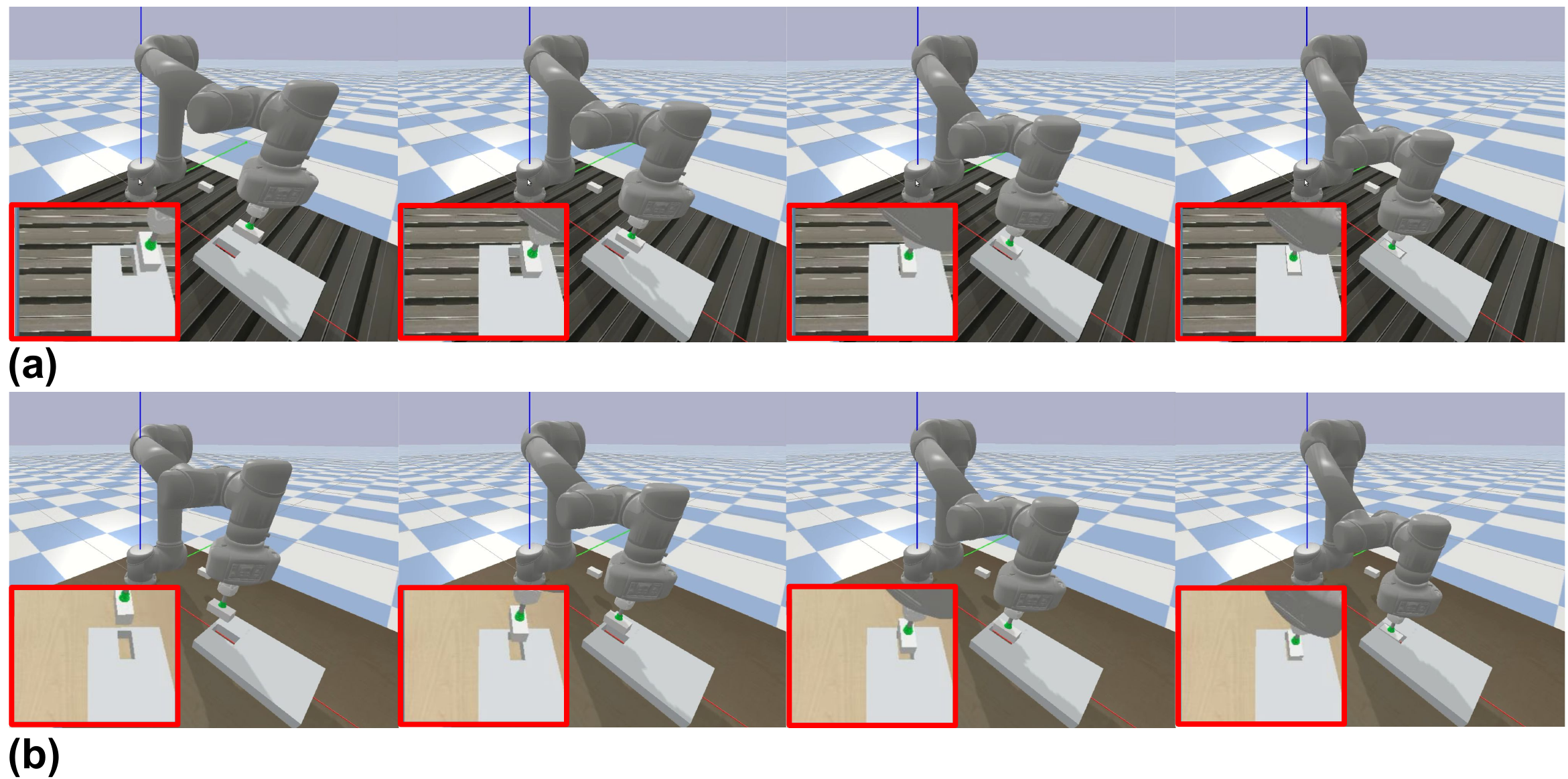}}
\caption{\textcolor{black}{(a) Execution phases of a scene of a white block with a metallic texture scene and (b) scene of a white block with a wooden texture from the initial pose to the target hole.}} 
\label{fig: execution phases}
\end{figure}
\begin{figure}[!h]
\centerline{\includegraphics[width=8.7cm,height=5.7cm]{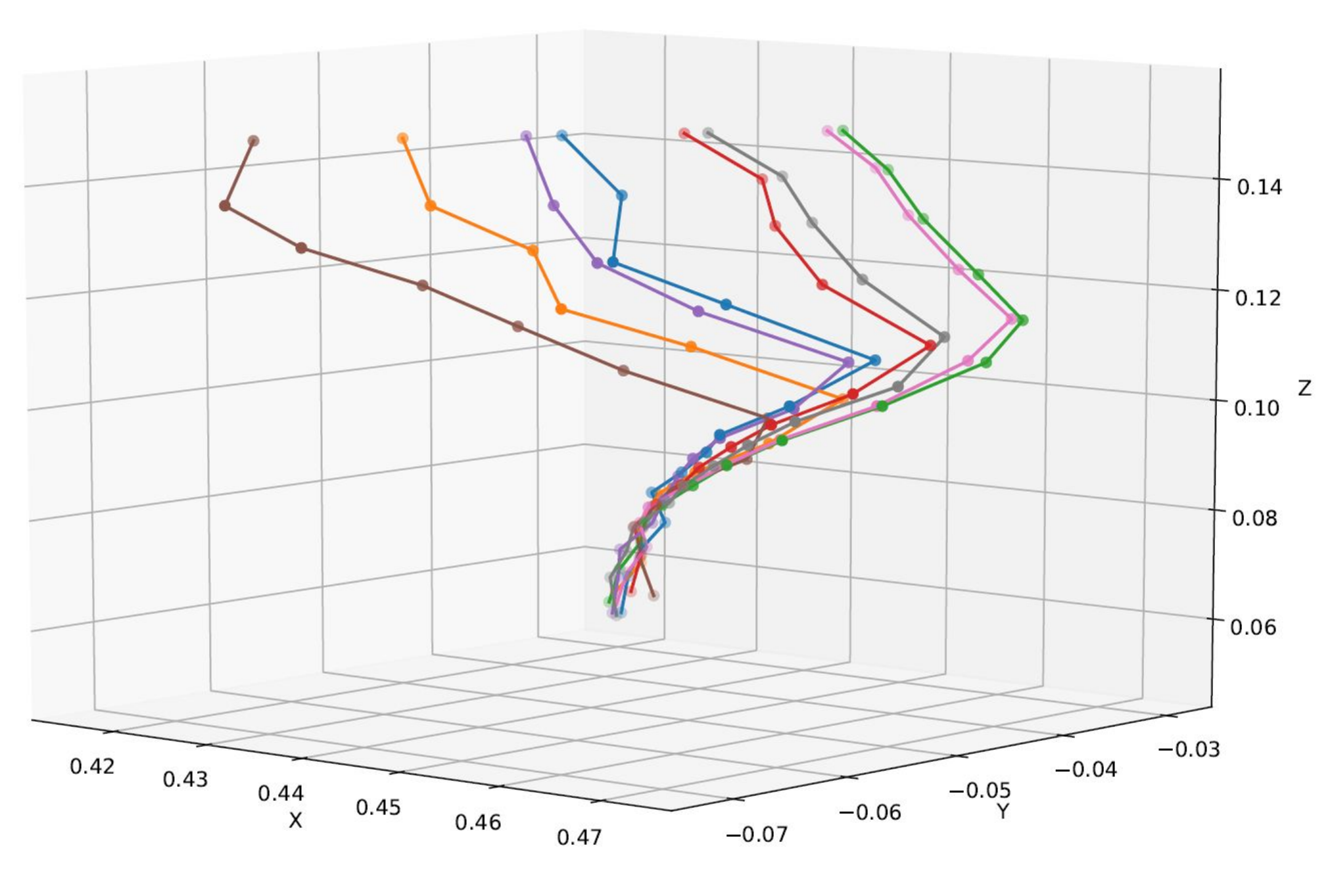}}
\caption{\textcolor{black}{Successful insertion trajectories starting from 8 different initial positions in the simulation. Eventually, the blocks are all moved into the target hole.}}
\label{fig:Successful insertion trajectories starting from 8 different initial positions}
\end{figure} In the simulation, we first generate a series of images of the robot state by executing random actions in the simulator as the training dataset and then train the autoencoder using this dataset. Thereafter, we extract the encoder as a part of the SAC network. We use the generate simulated images of the RGB observation space (size=$3\times 64\times 64$) to train the autoencoder.

We train the agent using cumulative episodes, and the results are shown in \Cref{tab:Performance with different observation space}. With the latent representation as policy input, the policy converged and the success rate could reach 96\% at checkpoints 3000 and 10000 episodes. However, the raw RGB and grayscale observation spaces cannot train a feasible policy, which is consistent with the results reported in a study \cite{schoettler2019deep}. 

We can achieve the answer to question 1 from \Cref{tab:Performance with different observation space}. We conclude that end-to-end learning is not as efficient as the approach that uses the latent representation observation space. Hence, we perform the remaining experiments using the latent representation observation space. 

\begin{table}[h!]
\caption{SUCCESS RATES OF DIFFERENT SCENES (SIMULATION)}
\label{tab:Performance in different scenes}
\centering
\begin{tabular}{@{}|c|c|c|@{}}
\toprule
\multicolumn{1}{|l|}{\textbf{Evaluate Trials}} & \textbf{Scene}                                      & \textbf{Success Rate} \\ \midrule
\multirow{4}{*}{500}                         & red block with wooden texture                       & 96\%                  \\ \cmidrule(l){2-3} 
                                              & red block with metal texture                        & 70.5\%                \\ \cmidrule(l){2-3} 
                                              & white block with wooden texture                     & 99\%                  \\ \cmidrule(l){2-3} 
                                              & \multicolumn{1}{l|}{white block with metal texture} & 96\%                  \\ \bottomrule
\end{tabular}
\end{table}
\begin{figure}[!h]
\centerline{\includegraphics[width=8.7cm,height=6.8cm]{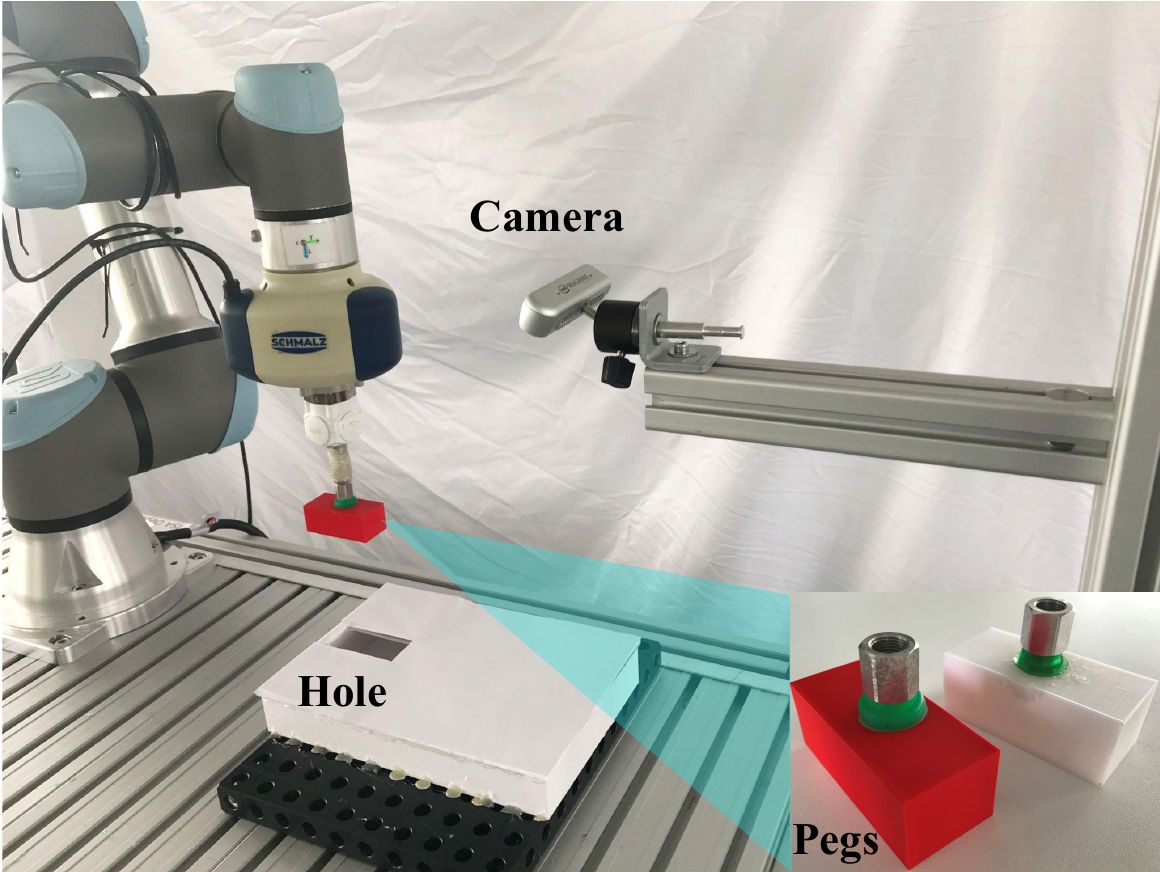}}
\caption{\textcolor{black}{Hardware settings for real environment experiment. The realsense D415 camera was installed manually without calibration, because we deliberately introduce uncertainties for the sim-to-real transfer to test it's robustness.}} 
\label{fig:Hardware settings}
\end{figure}

\subsubsection{Different Scene Evaluation}

Although we demonstrate the policy performance before using a latent representation observation space in the scene of a white block with a metallic texture, it is unclear whether the difference in the scene will influence the performance. A high success rate must be achieved in the simulation environment to perform further real experiments. Additionally, to verify the generalization of the framework, we consider the permutation of four environmental scenes with two blocks and two textures: a red block with a wooden texture, a white block with a wooden texture, a red block with a metallic texture, and  a white block with a metallic texture (\Cref{fig: fourscene}). Every scene is trained 3000 episodes in the simulation environment with a Dell Precision 5510 laptop CPU.

We compare the performance of this approach in different scenes, and the execution phase is presented in \Cref{fig: execution phases}. Additionally, a visualization of successful insertion trajectories is shown in \Cref{fig:Successful insertion trajectories starting from 8 different initial positions}. \Cref{tab:Performance in different scenes} shows the success rate of the framework obtained under different scenes. Three scenes achieved a success rate higher than 96\%. The white block with a wooden texture reached a 99\% success rate.

\subsection{Real-World Environment Experiment}

\begin{figure}[htbp]
\centerline{\includegraphics[width=8.8cm,height=5cm]{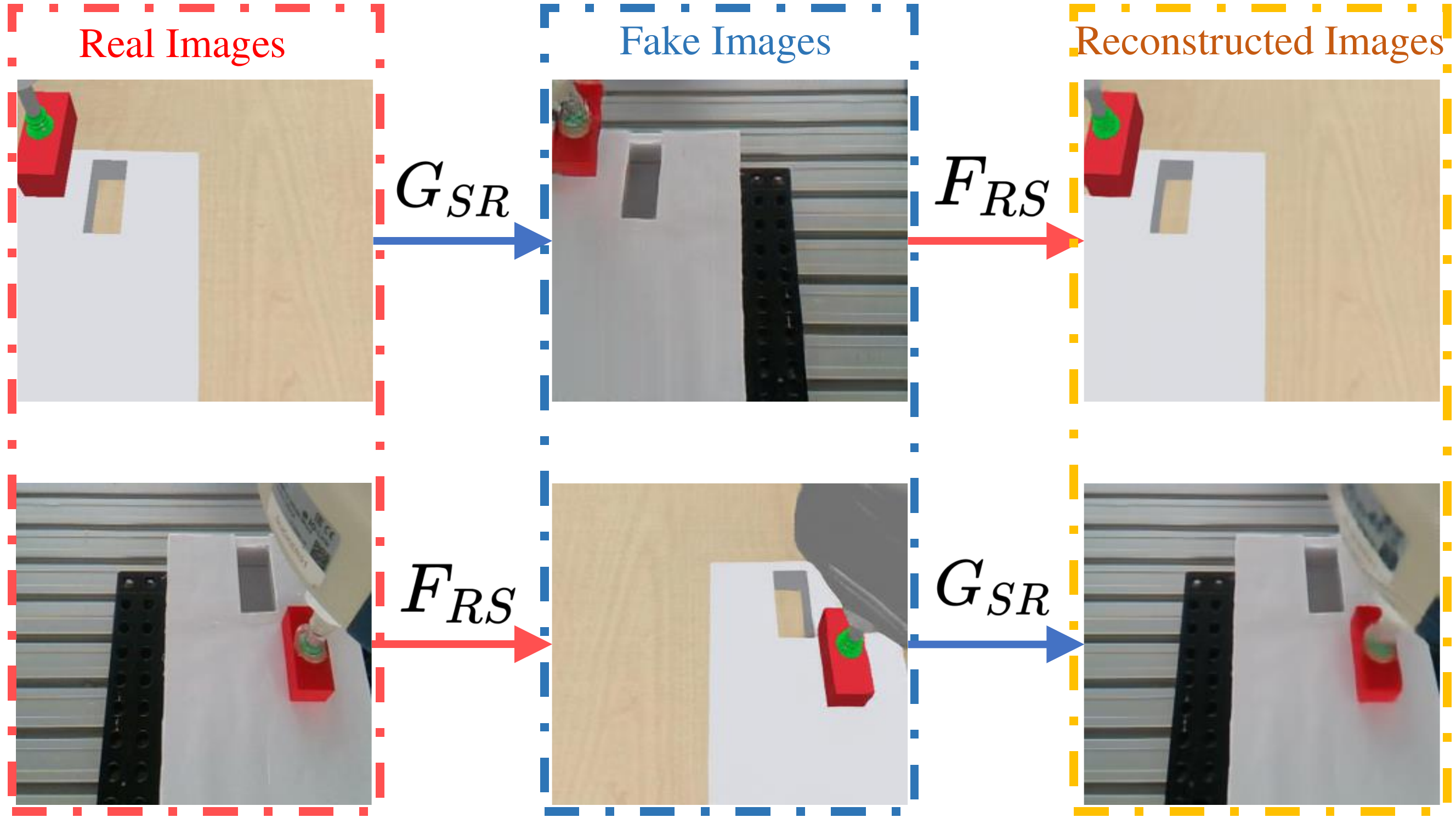}}
\caption{\textcolor{black}{Domain adaptation by CycleGAN: Mapping function $G_{SR}$, which maps the simulated distribution to a real world distribution. Mapping function $F_{RS}$, which maps the real world distribution to simulated distribution. }}
\label{fig:Domain adaptation by cycleGAN}
\end{figure}

Although we perform training and evaluate the success rate in the simulation environment, our goal is to transfer the trained policy to the physical environment. 

In our real-world environment setup (\Cref{fig:Hardware settings}), a UR5e robot\footnote{https://www.universal-robots.com/products/ur5-robot/} is used to perform a peg-in-hole insertion task. This 6-axis robot features a 5 kg payload and a working radius of 850 mm. It is equipped with a 6 degrees of freedom force/torque sensor on the end effector. The robot uses an operational space admittance controller \cite{hogan1985impedance} with 500 Hz control rate. The blocks are mounted behind the force/torque sensor to ensure the detection of the contact force with the environment.

An Intel RealSense D415 camera\footnote{https://www.intelrealsense.com/zh-hans/depth-camera-d415/} is fixed on the platform to observe the operation. The position and orientation of the camera are selected to ensure the block and hole are visible during most of the training time. \Cref{fig:Hardware settings} shows our hardware setup in the experiment. 
In our experiment, we use a white and a red block with same dimensions of $65 \times 30 \times 25$ mm, and a white block with a hole size of $70 \times 35 \times 30$mm. The clearance in each direction (i.e., length and width) is 5 mm. We select this setup because we aim to establish a potential scenario in which a packed data cable is inserted into a phone box in the mobile phone assembly line \cite{song2020robotic}. 

As described in \Cref{Transfer Framework to Real Environment}, the CycleGAN is introduced to perform the domain adaptation process to transfer the image distribution from the real world to the simulation.
We capture 200 images of the robot state in the real world and then generated a training dataset along with 3000 simulated images to train the CycleGAN. We train the CycleGAN model on four Nvidia 1080 Ti GPUs.  The domain adaptation results of the trained CycleGAN with our setup inputs is shown in \Cref{fig:Domain adaptation by cycleGAN}.

Based on the previous results listed in \Cref{tab:Performance in different scenes}, we can conclude that the scene with wooden texture achieve the highest success rate with our policy.
Hence, we transfer the real-world image to a scene of a block with a wooden texture using the DA method to evaluate our framework in a real-world setup. We define three situations when testing the policy in the real world as \cite{lee2019making}. \begin{figure}[!h]
\centerline{\includegraphics[width=8.5cm,height=2.2cm]{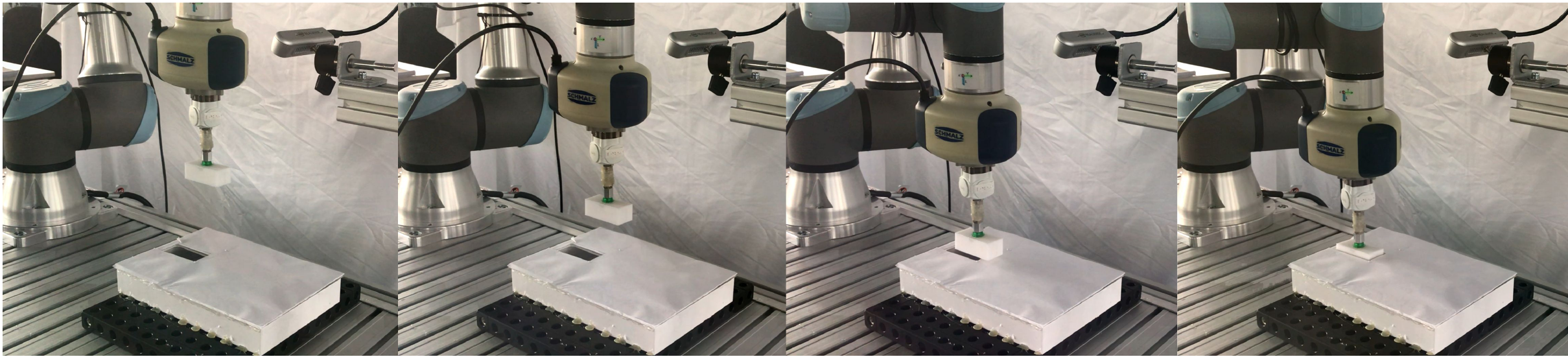}}
\caption{\textcolor{black}{White block with metal texture scene real setup execution phases from initial pose to target hole.}} 
\label{fig: real setup execution phases}
\end{figure}

\begin{table}[h!]
\caption{PERFORMANCES IN REAL ENVIRONMENT SETUPS}
\label{tab:Performance on real robot}
\centering
\begin{tabular}{@{}|c|c|c|c|@{}}
\toprule
\textbf{Scene} & \textbf{Complete insertion} & \textbf{Touched the box} & \textbf{Failed} \\ \midrule
Red block      & 86/100                      & 10/100                   & 4/100           \\ \midrule
White block    & 88/100                      & 12/100                   & 0/100           \\ \bottomrule
\end{tabular}
\end{table}
\textit{Complete Insertion} means that the robot accomplishes the insertion task completely. \textit{Touched the box} implies that the the peg was moved in the right direction, but the insertion is not completed. \textit{Failed} indicates a situation in witch the robot moves far away from the target in the wrong direction or performs unexpected movements.

During the execution (\Cref{fig: real setup execution phases}), we randomly occlude the camera's field of view for several seconds and push the robot in the wrong direction to the target hole to evaluate the system robustness to external perturbations. The performance of the physical robot in the real-world setup is summarized in \Cref{tab:Performance on real robot}. We obtain an average success rate equal to method reported in the literature\cite{lee2019making} with a safer sim-to-real framework because we limit the force command amplitude during the control.

%% file: chapters/06_conclusion.tex
\section{CONCLUSIONS AND DISCUSSIONS}\label{RESULTS AND DISCUSSION}

In this work, we proved that our sim-to-real framework is a valid approach to solving the peg-in-hole task both in simulated and real-world environments. By employing DA and force controller, we can directly transfer the policy that was trained in a simulator to a real-world setup.  
Moreover, we evaluated different observation spaces and proved that the latent representation (i.e., low dimension) can accelerate the convergence of policy learning and afford a higher success rate for the task than end-to-end learning using raw image input. The importance of force control is shown by the fact that in real-world experiments, the blocks often needs to contact the environment and ``slide'' into the hole.

However, our method can be optimized further in terms of the performance and generalization ability. For example, the  scene of a red block with a metallic texture in the simulation achieves a success rate of only 70.5\% considerably worse than those achieved using the other three scenes; hence, further research is needed.
Moreover, in the experiment setup, we assumed that the target orientation is known for the insertion task, which simplifies the task. For more general tasks, for example, when a robot starts with a random pose, our method must be improved to output the orientation of the end-effector. A more complex action space with both translation and rotation, $[\Delta x,\ \Delta y,\ \Delta z, \Delta r_{x}, \Delta r_{y}, \Delta r_{z} ]$, can be designed for training.
In our future roadmap, investigating the application of our sim-to-real approach to more industrial robotic tasks will be interesting.